\documentclass[twoside,leqno,twocolumn]{article}
\usepackage{ltexpprt}

\usepackage{xspace}
\usepackage{cite}
\usepackage{amsmath}
\usepackage{amsfonts, amssymb}
\usepackage{algorithm}
\usepackage{algpseudocode}
\usepackage{varwidth}
\usepackage{float}
\usepackage{framed}
\usepackage{array}
\usepackage{times}
\usepackage{color}
\usepackage{graphicx}
\usepackage{multirow}
\usepackage{pbox}
\usepackage{paralist}
\usepackage{listings}
\usepackage{url}
\usepackage{adjustbox}
\usepackage{booktabs}
\usepackage{caption} 
\usepackage{subcaption}
\DeclareMathOperator*{\argmax}{arg\,max}

\begin{document}
\newcommand{\hide}[1]{}
\newcommand{\atn}[1]{\textcolor{red}{#1}}
\newcommand{\notice}[1]{{\textsf{\textcolor{green}{{\em [#1]}}}}}
\newcommand{\reminder}[1]{{\textsf{\textcolor{blue}{[#1]}}}}
\newcommand{\vectornorm}[1]{\left|\left|#1\right|\right|}

\newcommand{\mkclean}{
    \renewcommand{\reminder}{\hide}
}

\newcommand{\bit}{\begin{compactitem}}
\newcommand{\eit}{\end{compactitem}}
\newcommand{\ben}{\begin{compactenum}}
\newcommand{\een}{\end{compactenum}}

\newtheorem{problem}{Problem}
\newtheorem{definition}{Definition}
\newcommand{\codeurl}{\url{www.andrew.cmu.edu/user/bhooi/ratings.tar}}
\newcommand{\model}{\textsc{BIRD}\xspace}
\newcommand{\metric}{\textsc{NEST}\xspace}
\newcommand{\method}{\textsc{BIRDNEST}\xspace}

\title{\method: Bayesian Inference for Ratings-Fraud Detection}
\author{Bryan Hooi\thanks{Carnegie Mellon University} \\
Neil Shah\footnotemark[1] \\
Alex Beutel\footnotemark[1] \\
Stephan G\"{u}nnemann\thanks{Technische Universit\"at M\"unchen} \\
\and
Leman Akoglu\thanks{Stony Brook University} \\
Mohit Kumar\thanks{Flipkart} \\
Disha Makhija\footnotemark[4] \\
Christos Faloutsos\footnotemark[1]
}
\date{}

\maketitle

\begin{abstract} \small\baselineskip=9pt
Review fraud is a pervasive problem in online commerce, in which fraudulent sellers write or purchase fake reviews to manipulate perception of their products and services. Fake reviews are often detected based on several signs, including 1) they occur in short bursts of time; 2) fraudulent user accounts have skewed rating distributions. However, these may both be true in any given dataset. Hence, in this paper, we propose an approach for detecting fraudulent reviews which combines these 2 approaches in a principled manner, allowing successful detection even when one of these signs is not present. To combine these 2 approaches, we formulate our Bayesian Inference for Rating Data (\model) model, a flexible Bayesian model of user rating behavior. Based on our model we formulate a likelihood-based suspiciousness metric, Normalized Expected Surprise Total (\metric). We propose a linear-time algorithm for performing Bayesian inference using our model and computing the metric. Experiments on real data show that \method successfully spots review fraud in large, real-world graphs: the $50$ most suspicious users of the Flipkart platform flagged by our algorithm were investigated and all identified as fraudulent by domain experts at Flipkart. 
\end{abstract}

\section{Introduction}
\label{sec:Introduction}

\begin{figure*}[!htp]
    \centering
    \begin{subfigure}[t]{0.49\textwidth}
        \centering
	    \includegraphics[width=\textwidth]{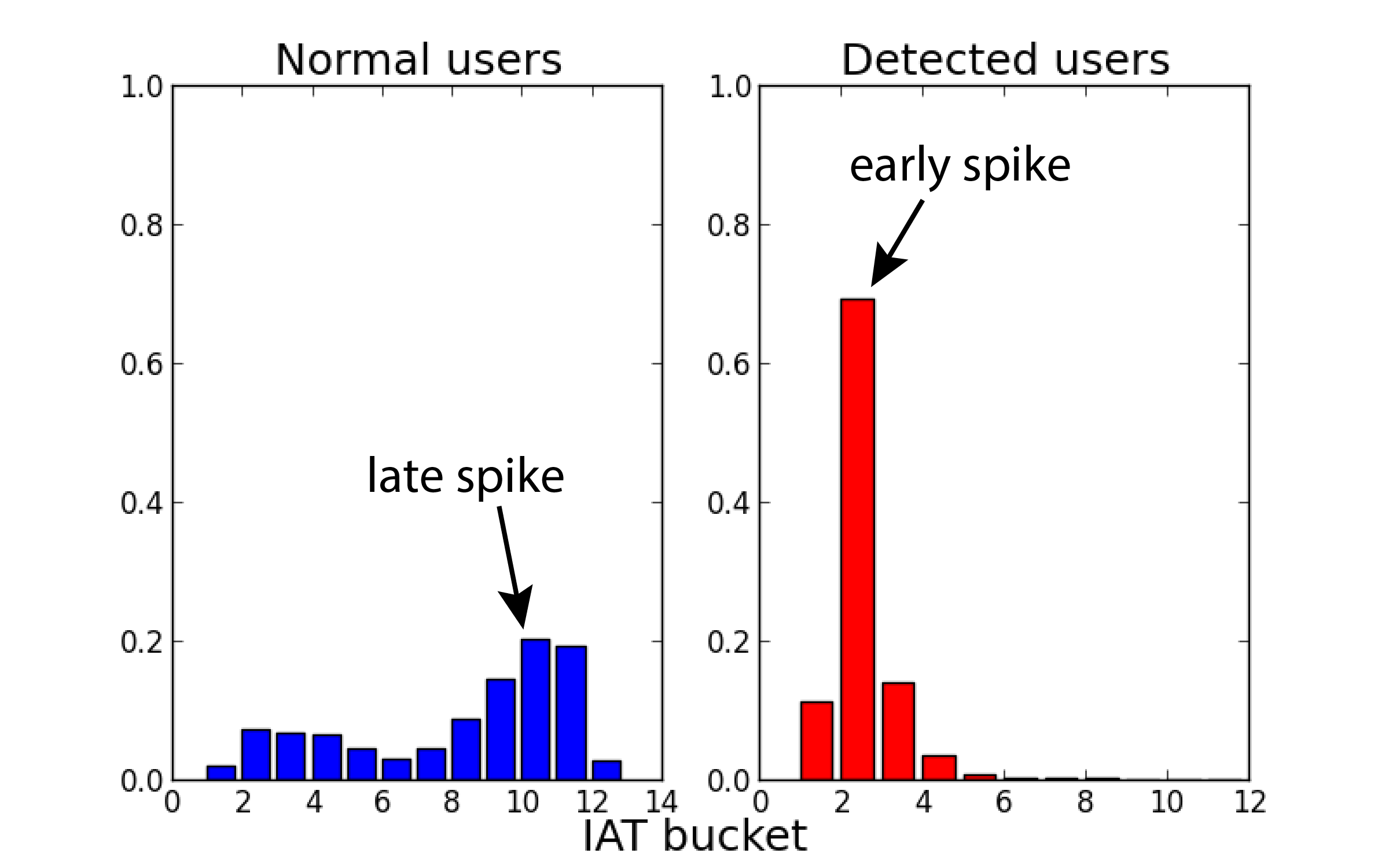}
	    \caption{\textbf{Common pattern observed that detected users' ratings are more `bursty' than normal users.}. Times between a user's ratings were bucketed logarithmically; detected users have shorter times between ratings.}
	\label{fig:flipkart_iat_goodbad}
    \end{subfigure}
    \hfill
    \begin{subfigure}[t]{0.49\textwidth}
        \centering
    \includegraphics[width=0.9\textwidth]{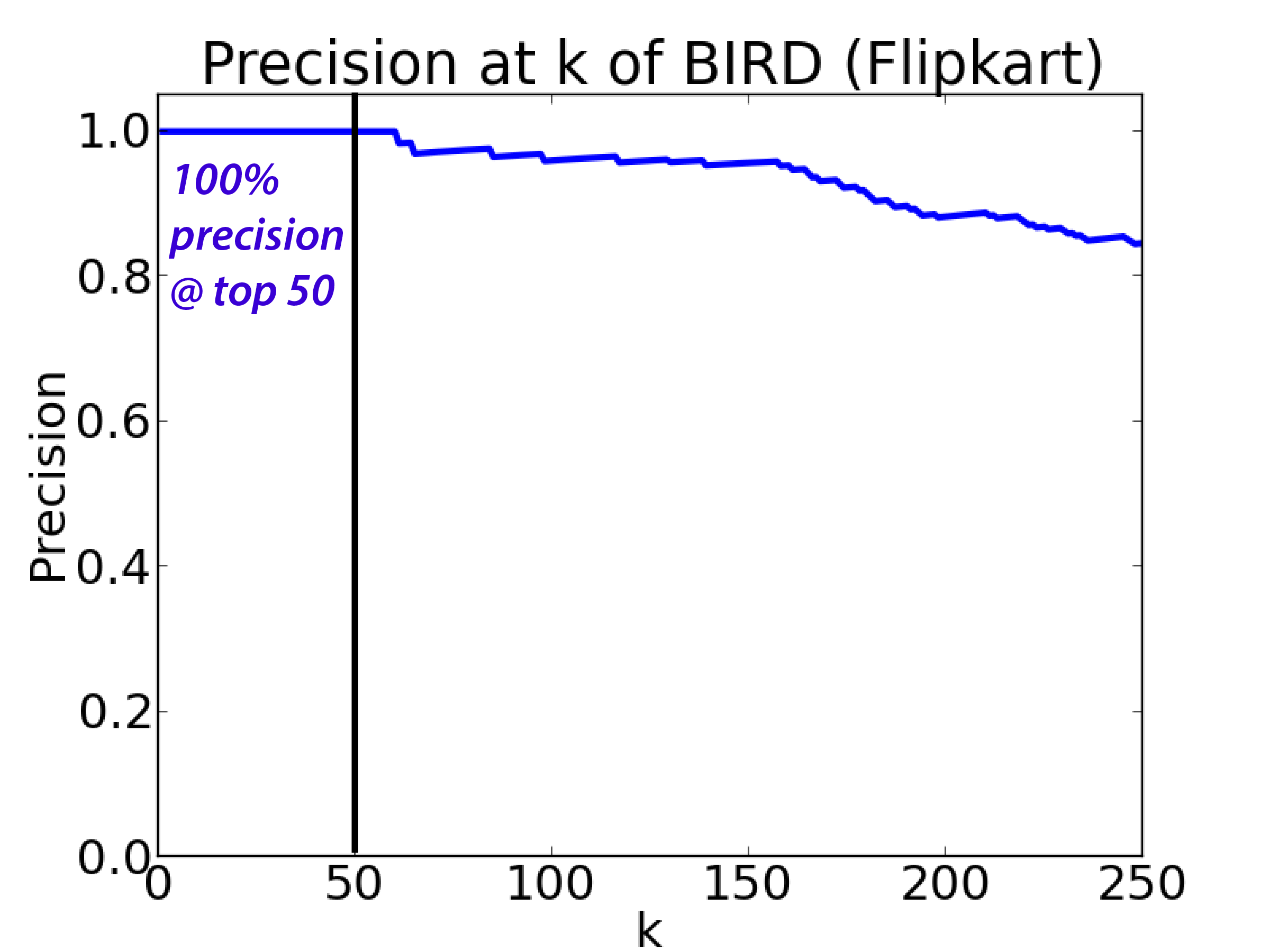}
	    \caption{\textbf{\textsc{BIRDNEST} is effective in practice}, with $211$ users of the top $250$ flagged by \textsc{BIRDNEST} involved in fraud.}
		\label{fig:flipkart_pak}
    \end{subfigure}
    \hfill
    \caption{\label{fig:cj}{\textbf{\textsc{BIRDNEST} combines temporal and rating information in a principled manner to detect fraud with high precision.} Inspecting the most suspicious $100$ users shows their strongly anomalous patterns.}}
\end{figure*}

Online reviews play an important role in informing customers' purchasing decisions. This has led to the problem of fake reviews, in which businesses write or purchase fake reviews in order to raise the popularity of their products or services. Hence, it is crucial for online commercial platforms to identify and remove these reviews, in order to maintain customers' trust in the accuracy of their reviews.

Various inputs such as rating, review text, timestamp etc. may be available for detection systems; in this work we focus on ratings and timestamps as they are commonly available and informative features. Informally, our problem is:

\begin{problem}[Informal]
\textbf{Given} a set of users and products, and timestamped ratings (e.g. $1$ to $5$ stars) by users for products, \textbf{compute } a suspiciousness score for each user.
\end{problem}

Currently, a number of algorithms use a temporal approach to detect ratings fraud \cite{ferraz2015rsc,gunnemann2014detecting,xie2012review}. These focus on catching products that receive a large number of positive or negative reviews in a short time, motivated by the `bursty' nature of fraudulent reviews when a store wishes to rapidly increase their popularity or defame their competitors. An alternative approach based on rating distributions is to focus on finding users who rate products very differently from other users \cite{lim2010detecting,jindal2010finding}. These focus on detection of suspicious behavior by users or products in terms of their deviation from normal practice. 

In this paper, we aim to combine both approaches in a principled way by constructing a Bayesian model for rating behavior, then formulating a likelihood-based metric which measures how much a user deviates from the rest of the users. 

The Bayesian approach also provides a principled solution to the conceptually difficult problem of finding a good tradeoff between users with extreme rating distributions vs. users with larger number of ratings. Is a user with $50$ ratings (average rating $5.0$) more suspicious than a user with $500$ ratings (average rating $4.95$)? Bayesian methods allow us to quantitatively answer this question. Namely, our Bayesian method combines the rating distribution and number of ratings to estimate our \emph{beliefs} about the rating characteristics of a user in a way that captures our uncertainty, which then determines how suspicious the user is.

Our contributions are:

\bit
\item \textbf{Theoretically sound user behavior model}: we define a Bayesian model for the data based on a mixture model which captures different types of user behavior. This model then allows us to determine how much an anomalous user deviates from normal behavior.
\item \textbf{Suspiciousness metric}: we define a likelihood-based metric which measures how much a user deviates from normal behavior.
\item \textbf{Algorithm}: we propose a scalable and effective algorithm for learning the Bayesian model and evaluating suspiciousness.
\item \textbf{Effectiveness}: we show that our method successfully spots review fraud in large, real-world graphs, with precision of over $84\%$ on the top $250$ Flipkart users flagged by our algorithm.
\eit

\begin{figure}[t]
    \centering
    \includegraphics[width=0.5\textwidth]{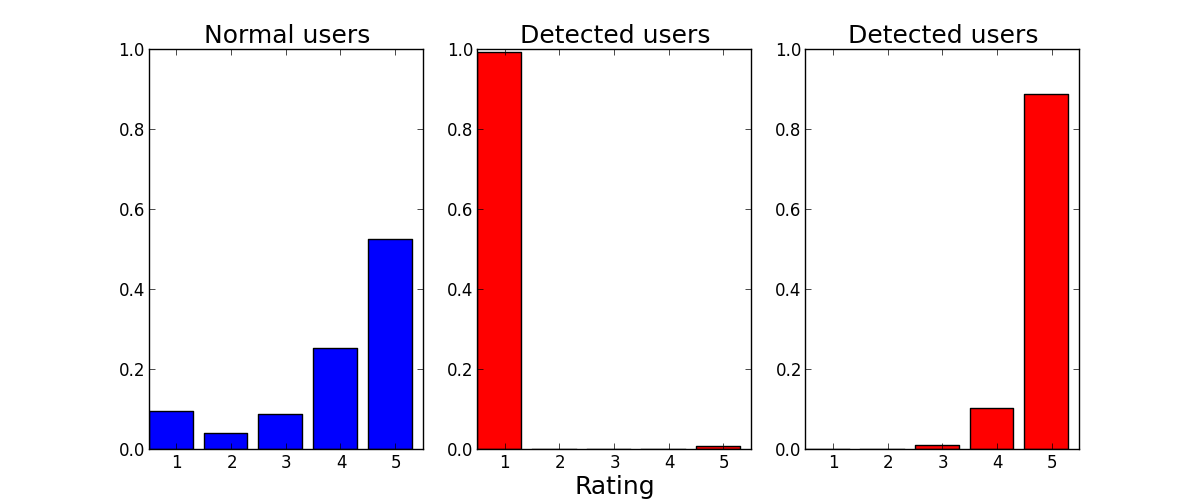}
    \caption{\textbf{Common pattern observed that detected users' ratings deviate strongly from normal users}: inspecting the detected users shows that they consists of two groups: highly negative users (middle) and highly positive users (right).}
\label{fig:flipkart_ratings_goodbad}
\end{figure}

\textbf{Reproducibility:} our code is open-sourced at \codeurl.

\section{Background and Related Work}
\label{sec:background}

\subsection*{Content-based approaches}
A significant portion of opinion fraud comes from customer reviews online.
Customer reviews have been long studied \cite{hu2004mining}, and 
many methods for review fraud focus on review text, such as
\cite{ott2011finding,jindal2008opinion,feng2012syntactic}. 
While these methods are illuminating, many sites only have ratings without
text, or text is easily manipulated.
Therefore, in our setting, we focus on ratings and their temporal
characteristics, as review text is not always available.

\subsection*{Graph-based approaches}
Much of the existing work in fraud or anomaly detection on graphs has focused on
detecting fraud in pure graphs; that is, graphs with no node or edge labels.
This includes spectral methods which use eigen-decomposition or singular value
decomposition (SVD) to group similar nodes in the graph
\cite{prakash2010eigenspokes,jiang2014inferring,shah2014spotting}.
\cite{wang2011review} uses an iterative approach to label as honest and
dishonest. Approaches based on Markov Random Fields and belief propagation have
also been used to identify dense or suspicious subgraphs \cite{pandit2007netprobe,akoglu2013opinion}.
\cite{ye2015discovering} detects spammers through
graph-based measures measuring self-similarity and neighborhood diversity.
However, these methods do not make use of key temporal and rating data. 

\subsection*{Temporal methods for fraud detection}
There are a number of works on anomaly detection in multivariate time series
\cite{li2007mining,
cheng2009detection,vahdatpour2010unsupervised,ramaswamy2000efficient}. 
\cite{beutel2013copycatch} focuses on fraudulent temporal patterns in graphs,
and \cite{ferraz2015rsc} found suspicious inter-arrival times between events in
social media.
A couple of works address temporal patterns of reviews, e.g.
\cite{xie2012review} detects spam singleton reviews and
\cite{gunnemann2014detecting} detects time periods of unusual activity.
However, our goal is to compute a general, principled, likelihood-based
measurement of how suspicious each user is. 
In this regard, \cite{crosspot} offers a general suspiciousness metric for
count data but is not suitable for ratings data.

\subsection*{Behavior modeling and fraud detection}
A wide body of research has focused on understanding user behavior and
especially rating behavior.  In particular ratings have been studied by the
recommendation systems community, with both frequentist
\cite{koren2008factorization} and Bayesian models
\cite{salakhutdinov2008bayesian} demonstrating great success.
Additionally some models have worked to take into account temporal features
\cite{koren2010collaborative}, and others have captured the bimodal patterns in
ratings data \cite{beutel2014cobafi}.

Other behavior models have been proposed to detect users who
deviate from normal practice in a meaningful way \cite{lim2010detecting,jindal2010finding}. 
In \cite{edgecentric} a similar problem of finding anomalies in temporal rating data was treated with information theoretic arguments.  By taking a Bayesian approach, we develop a significantly different perspective on the problem and our resulting metric of suspiciousness is more flexible, allowing for explicit priors, unique posteriors for each user, and easy extensions to other distributions.

\label{sec:Method}
\section{Bayesian Model}

\subsection{Motivating Example}
\label{sec:motivating}
\begin{figure}[htp]
    \centering
    \includegraphics[width=0.5\textwidth]{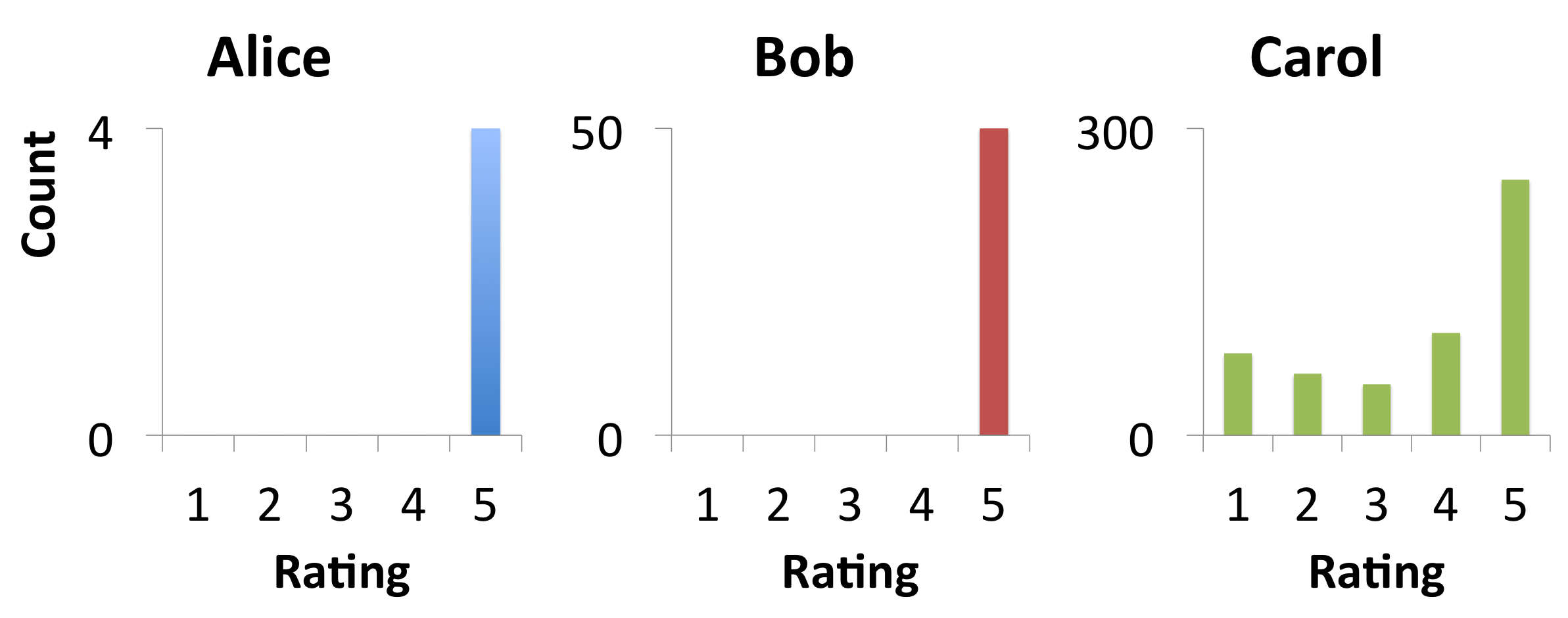}
    \caption{Rating distributions of example users. The histograms show how many times each user gave each star rating.}
\label{fig:example_users}
\end{figure}

We start by illustrating why a Bayesian approach is helpful. Consider users Alice, Bob and Carol whose rating distributions are as given in Figure \ref{fig:example_users}. For example, Alice rated $4$ products, all with $5$ stars. Bob did the same, $50$ times. Carol gave about $300$ ratings, and exhibits a `hockey-stick' distribution, which is close to the average over all users. Which user is the most suspicious (i.e. likely fraudulent)? Our goal is to come up with a principled and intuitive measure of how suspicious each user is.

Why does Alice's low rating count makes her less suspicious than Bob? Our answer is: since we only have $4$ products rated by Alice, we have little information about her true (i.e. long-term) rating behavior. She may simply be a normal user who appears unusual as her first few ratings were high, but given more ratings, she would converge to a more typical distribution. Bob, however, is much less likely to be a normal user: we can say with greater certainty that his true rating behavior is anomalous. 

Intuitively, deciding how suspicious each user is involves a two-step process: first, we estimate our beliefs for what that user's true rating distribution is. Second, we estimate how suspicious we believe they are, given our beliefs. For Alice, our beliefs are highly uncertain: we cannot be confident that her rating distribution is unusual. For Bob, we are confident that his rating distribution is fairly skewed toward 5s. For Carol, we know her rating distribution with high confidence, but it is not suspicious. 

\begin{figure}[htp]
    \centering
    \includegraphics[width=0.4\textwidth]{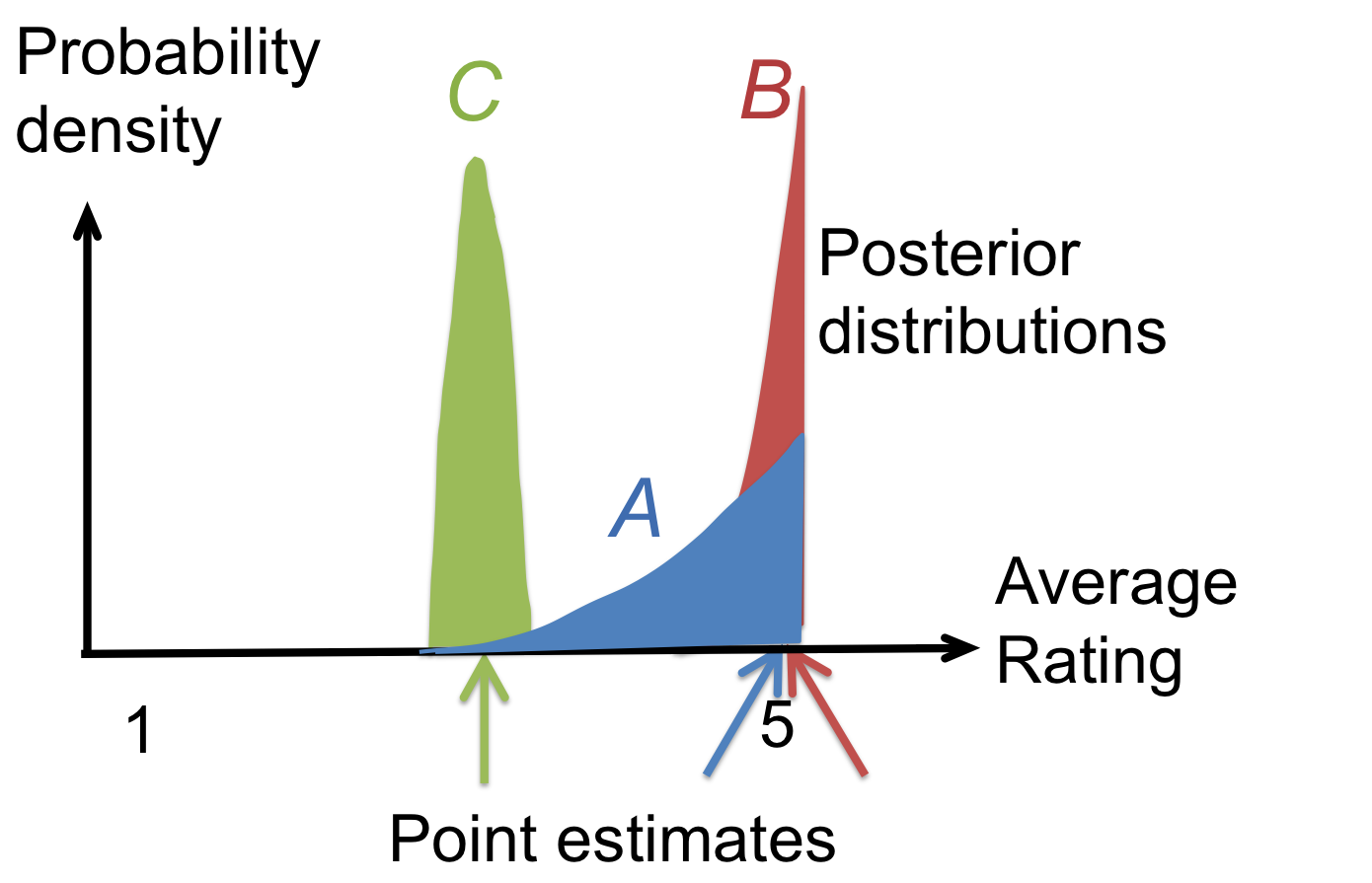}
    \caption{\textbf{Posterior distributions, not point estimates, mark a user as suspicious.} Bob is suspicious because our beliefs about his true average rating are both narrow and close to 5, while Alice is less suspicious because our beliefs about her true average rating are more spread out. }
\label{fig:posterior_distributions}
\end{figure}

The Bayesian approach applies this intuition in a principled manner. It first sets a prior, estimated from data, representing our `default' beliefs about users's rating behavior. It then estimates our beliefs (in the form of a posterior distribution) about their rating distribution. Finally, we compute how suspicious we believe them to be, averaging over their posterior distribution. Figure \ref{fig:posterior_distributions} illustrates how posterior distributions capture the information we need to identify a user as suspicious. The posterior distributions in Figure \ref{fig:posterior_distributions} refer to our beliefs about each user's true long-term average rating, expressed as a probability distribution. The point estimates refer to each user's observed average rating, which do not capture how much more certain we are in Bob's case than Alice, and hence how much more suspicious Bob is. 

\paragraph{Alternatives that don't work ($z$ or $t$ tests)} What about instead performing a standard hypothesis test (such as a $z$ or $t$-test) for each user's average rating (or any other quantity associated with their rating distribution), to see whether their average rating differs significantly from the population? The problem with this approach lies with users like Carol, who differ slightly from the population but have a large number of ratings. Even as normal (non-fraudulent) users, we expect their true average rating to differ slightly from that of the population (say, by $0.1$) just due to inter-person variation. 

Given enough ratings, however, even such a small difference could produce arbitrarily small p-values under such a hypothesis test, since the test correctly concludes that there is an extremely small probability of drawing Carol's observed average rating if her true average rating were equal to that of the population. However, such small differences are not suspicious. The Bayesian approach would instead estimate Carol's posterior distribution as in Figure \ref{fig:posterior_distributions} and conclude that it is both narrow and entirely non-suspicious, which is a more sensible result. 

\subsection{Proposed Model}

Table \ref{tab:symbols} summarizes the notation used in this paper.

\begin{table}[htp]
\centering
\caption{Commonly used notation in this paper. Vectors are in \textbf{bold}.}
\label{tab:symbols}
\begin{tabular}{@{}rl@{}}
\toprule
\textbf{Parameter}           & \textbf{Interpretation}                                                   \\ \midrule
\textit{$m$}                 & No. of users                                                           \\
\textit{$n_i$}               & No. of ratings given by user $i$                                       \\
\textit{$s$}                 & No. of star levels (e.g. $s=5$ for $1$ to $5$ stars)                   \\
\textit{$x_{ij}$}            & Rating of the $j$th rating given by user $i$                   \\
\textit{$b$}            & base of logarithm for temporal bucketing                   \\
\textit{$\Delta_{ij}$}       & temporal bucket of the $j$th rating by user $i$                          \\
\textit{$\Delta_{max}$}      & temporal bucket with highest index                                             \\
\textit{$\mathbf{x_i}, \boldsymbol{\Delta_i}$}      & Vector $(x_{ij})_{j=1}^{n_i}$ (resp. $\Delta_{ij})_{j=1}^{n_i}$)  		\\
\textit{$X, \Delta$}      & Matrix containing all the $(x_{ij})$ (resp. $(\Delta_{ij})$)      \\
\textit{$n^x_{il},n^\Delta_{il}$}   & No. of times user $i$ gave rating (resp. time) $l$   \\
\textit{$\mathbf{n}^x_i,\mathbf{n}^\Delta_i$}   & Vector $((n^x_{i1}),\dots,(n^x_{is}))$ (resp. $((n^\Delta_{i1}),\dots))$  \\
\textit{$K$}                 & No. of clusters                                                        \\
\textit{$\pi_k$}             & Probability of a random user being in cluster $k$                         \\
\textit{$z_i$}               & Cluster (or mixture component) of user $i$                                \\
\textit{$\mathbf{p_i}, \mathbf{q_i}$}          & User $i$'s rating (resp. temporal) distribution            \\
\textit{$\boldsymbol{\alpha_k}, \boldsymbol{\beta_k}$}          & Dirichlet parameters for cluster $k$          \\
$F_x, F_\Delta$ & Global distributions; refer to \eqref{eq:global_distribution} \\ 
\bottomrule
\end{tabular}
\end{table}

In our problem setting, users are indexed $i=1,\dots,m$. User $i$ has $n_i$ ratings, indexed by $j=1,\dots,n_i$. The ratings in stars given by user $i$ are denoted by the variables $x_{ij} \in \{1, 2, \dots, s\}$ (e.g. for star ratings from $1$ to $5$ we have $s=5$). Similarly to \cite{ferraz2015rsc}, we preprocess the rating timestamps by computing its time difference from the previous rating, i.e. the difference between its timestamp and the timestamp of the last rating given by the same user. We then bucket the time differences according to the integer part of the log base $b$, where $b$ is chosen to result in close to $20$ buckets. The temporal bucket of the $j$th rating of user $i$ is denoted $\Delta_{ij} \in \{1, 2, \dots, \Delta_{max}\}$ for $j=1,\dots,n_i$, analogous to $x_{ij}$.

Using time differences instead of raw timestamps makes it possible to detect either unusually rapid rating of products by a user (due to having a concentration of small time differences), or unusually regular patterns, such as rating products once every hour. Both of these patterns suggest bot-like or spammy behavior, which we would like to detect. Moreover, the discretized i.e. multinomial approach allows us to flexibly detect a wide range of possible deviations from normal behavior without assuming a more restrictive parametric form, such as a Gaussian distribution.

We will consider the ratings $X$ and time differences $\Delta$ to be generated based on a model. From a high level, our generative model for user behavior is a mixture model in which each user belongs to one of $K$ clusters: in general, there is no single type of user behavior, so we use clusters to capture different types of user behavior. Each cluster represents a certain type of rating distribution and temporal distribution for the users in that cluster.

Let $k=1,\dots,K$ index into the $K$ clusters. For each user $i$, we first generate which cluster they belong to, $z_i \in \{1, 2, \dots, K\}$, from a Multinomial($\boldsymbol{\pi}$) distribution, where $\pi_k$, the $k$th entry of $\boldsymbol{\pi}$, is the probability that a random user is generated in cluster $k$.

Even within a single cluster, it would not be reasonable to expect all users to behave exactly the same way. Thus, instead of using a single rating/temporal distribution per cluster, we allow small deviations per user. We do this by associating a common Dirichlet prior with each cluster: each user has their individual rating distribution drawn from this prior. We denote user $i$'s rating distribution by $\mathbf{p_i}$, a vector of length $s$ of nonnegative entries which sums to $1$, where the $j$th entry of this vector gives their probability of giving the $j$th rating. Thus, we draw user $i$'s rating distribution $\mathbf{p_i}$ from a Dirichlet($\boldsymbol{\alpha}_{z_i}$). Similarly, $\mathbf{q_i}$ represents user $i$'s temporal distribution, and we draw $\mathbf{q_i} \sim $ Dirichlet$(\boldsymbol{\beta}_{z_i})$.

Finally, to generate user $i$'s ratings, we draw each rating $x_{ij}$ based on user $i$'s rating distribution: $x_{ij} \sim $ Multinomial($\mathbf{p_i}$). Similarly, for the temporal buckets, we draw each $\Delta_{ij}$ from a Multinomial($\mathbf{q_i}$) distribution. 

The generative model we have described is summarized in \eqref{eq:generative_model}.

\begin{align}
z_i \ & \sim \ \text{Discrete}(\boldsymbol{\pi}) \nonumber \\
\mathbf{p_i} | z_i = k \ & \sim \ \text{Dirichlet}(\boldsymbol{\alpha}_k) \nonumber \\
x_{ij} \ & \sim \ \text{Multinomial}(n_i, \mathbf{p_i}) \nonumber \\
\mathbf{q_i} | z_i = k \ & \sim \ \text{Dirichlet}(\boldsymbol{\beta}_k) \nonumber \\
\Delta_{ij} \ & \sim \ \text{Multinomial}(n_i, \mathbf{q_i}) \label{eq:generative_model}
\end{align}

\begin{figure}[ht!]
\centering
 \includegraphics[width=0.4\textwidth]{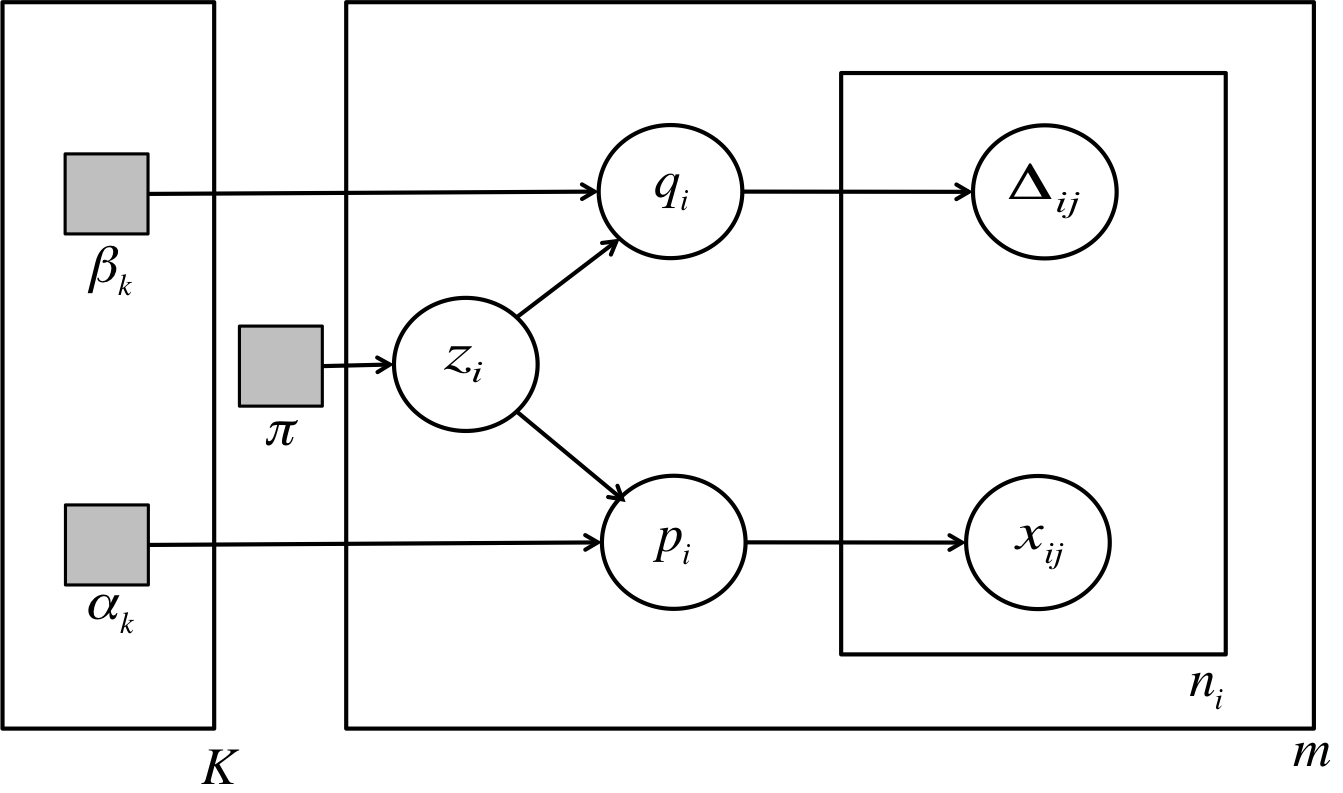}
 \caption{Graphical model describing users, ratings and rating times. User $i$'s mixture component $z_i$ determines how we generate their individual multinomial parameter vectors $p_i, q_i$, which then generate $x_{ij}$ and $\Delta_{ij}$ as samples from these multinomial distributions.}
 \label{fig:rating_graphical_model}
\end{figure}

The corresponding graphical model is given in Figure \ref{fig:rating_graphical_model}. 

\section{Proposed Algorithms}

\subsection{Fitting our Bayesian Model (\textsc{BIRD})}

Algorithm \ref{alg:fitting} fits data to the model of Fig.\ref{fig:rating_graphical_model}, by using a greedy hill climbing approach to maximize the overall likelihood function. In this algorithm, we iteratively adjust each parameter and the cluster assignments $z$ until convergence. Each of the $\arg\max$ lines in the algorithm can be solved efficiently, which we next describe how to do. 

\subsubsection*{Cluster parameters}\hfill \\
Here we fix $z$ and compute $\argmax_{\boldsymbol{\alpha}_k} P(X,\Delta | \boldsymbol{\alpha}_k, z)$ in Line \ref{line:adjust_alpha}; adjusting with respect to $\boldsymbol{\beta}$ will be similar. Note that adjusting $\boldsymbol{\alpha}_k$ only affects the likelihood with respect to $x_i$, for $i$ in cluster $k$. Thus we are equivalently maximizing $\prod_{i: z_i=k} P(x_i | \boldsymbol{\alpha}_k, z)$.

To be clear, here $P(x_i | \boldsymbol{\alpha}_k, z)$ refers to the marginal likelihood, i.e. the probability of generating $x_i$, after marginalizing out $\mathbf{p_i}$. Thus we need to find the maximum likelihood update for $\boldsymbol{\alpha}_k$ given the $x_i$ for $i$ in cluster $k$, which were sampled from the two-step process of first generating $\mathbf{p_i} \sim \text{Dirichlet}(\boldsymbol{\alpha}_k)$ and then generating $x_i \sim \text{Multinomial}(\mathbf{p_i})$. This two-step process is also known as the Dirichlet-multinomial distribution; \cite{minka2000estimating} provide fixed-point iteration methods for maximum likelihood estimation of $\boldsymbol{\alpha}_k$ in this setting. Specifically, we repeat until convergence, for each $k=1,\dots,K$ and $l=1,\dots,s$:

\begin{align}
 \alpha_{kl}^{new} &= \alpha_{kl} \frac{\sum_{i=1}^m \frac{n^x_{il}}{n^x_{il} - 1 + \alpha_{kl}}}{\sum_{i=1}^m \frac{n^x_{i}}{n^x_{i} - 1 + \sum_{l'} \alpha_{kl'}}}
 \label{eq:adjust_alpha}
\end{align}
Similarly, the update for $\beta$ is:
\begin{align}
 \beta_{kl}^{new} &= \beta_{kl} \frac{\sum_{i=1}^m \frac{n^\Delta_{il}}{n^\Delta_{il} - 1 + \beta_{kl}}}{\sum_{i=1}^m \frac{n^\Delta_{i}}{n^\Delta_{i} - 1 + \sum_{l'} \beta_{kl'}}}
 \label{eq:adjust_beta}
\end{align}

\subsubsection*{Cluster assignments} In Line \ref{line:adjust_z}, we fix the cluster parameters and fit the maximum likelihood cluster assignment $z_i$. Note that changing $z_i$ only affects the likelihood with respect to user $i$. Referring to our graphical model in Figure \ref{fig:rating_graphical_model}, maximizing $P(X,\Delta | z_i = k)$ is equivalent to finding:
\begin{align}
z_i = \arg\max_{k} \pi_k P(\mathbf{x_i} | z_i = k) P(\boldsymbol{\Delta_i} | z_i = k)
\label{eq:cluster_assignment}
\end{align}
To compute $P(\mathbf{x_i} | z_i = k)$, note that this is the probability of drawing $\mathbf{x_i}$ from a Dirichlet-multinomial distribution with known parameter $\boldsymbol{\alpha}_k$. 

Let $n^x_{il} = \sum_{j=1}^{n_i} 1\{ x_{ij}=l \}$ be the number of user $i$'s ratings that equal $l$, and similarly $n^\Delta_{il} = \sum_{j=1}^{n_i} 1\{ \Delta_{ij}=l \}$. The marginal distribution of a Dirichlet-multinomial distribution (after marginalizing out $\mathbf{p_i}$) is known to be
\begin{align*}
P(x_i | z_i = k) = \frac{\Gamma(A_k)}{\Gamma(n_i + A_k)} \prod_{l=1}^s \frac{\Gamma(n^x_{il} + \alpha_{kl})}{\Gamma(\alpha_{kl})}
\end{align*}
where $\Gamma$ is the gamma function, and $A_k = \sum_l \alpha_{kl}$. The term $P(\boldsymbol{\Delta_i} | z_i = k)$ can be computed in the same manner. Since $z_i$ is discrete, we can thus maximize \eqref{eq:cluster_assignment} by computing $\pi_k P(\mathbf{x_i} | z_i = k) P(\boldsymbol{\Delta_i} | z_i = k)$ for each value of $k$ and choosing the maximizing value of $k$.

\subsubsection*{Posterior distributions of $p$ and $q$} Here we explain how to compute the posterior distributions in Line \ref{line:posterior} of Algorithm \ref{alg:fitting}. Let $\mathbf{n}^x_i = ((n^x_{i1}),\dots,(n^x_{is}))$ and $\mathbf{n}^\Delta_i = ((n^\Delta_{i1}),\dots,(n^\Delta_{is}))$. At this point the entire iterative process of estimating the hyperparameters and cluster assignments is complete, and we have to compute the posterior distributions of $\mathbf{p_i}$ and $\mathbf{q_i}$ given the data $X$ and $\Delta$. $\mathbf{p_i}$ has a $\text{Dirichlet}(\boldsymbol{\alpha}_{z_i})$ prior, so by the conjugate prior property of Dirichlet distributions, its posterior distribution is $\text{Dirichlet}(\alpha_{z_{i}} + \mathbf{n}^x_i)$. Similarly, the posterior distribution of $\beta$ is $\text{Dirichlet}(\beta_{z_{i}} + \mathbf{n}^\Delta_i)$. 

\subsubsection*{Number of clusters} We select the number of clusters $K$ using the Bayesian Information Criterion (BIC). 

\subsubsection*{Convergence} As we can see from Algorithm \ref{alg:fitting}, each adjustment to $\boldsymbol{\pi}, \alpha, \beta$ or $z$ is an $\arg\max$ step and increases the overall likelihood $P(X,\Delta|z;\boldsymbol{\pi}, \alpha, \beta)$. Because the overall likelihood is bounded, this must converge.

\begin{algorithm}[t]
\caption{Fitting parameters for the model in Figure \ref{fig:rating_graphical_model}. $X$ is a matrix containing all the $x_{ij}$ and $\Delta$ is a matrix containing all the $\Delta_{ij}$.}
\label{alg:fitting}
\begin{algorithmic}[1]
\Procedure{\model}{$X, \Delta$}
	\State{\textbf{Output:} 
		\State cluster hyperparameters $\boldsymbol{\pi}, (\boldsymbol{\alpha}_k, \boldsymbol{\beta}_k)_{k=1}^K$
		\State posterior distributions for each user's rating and temporal distribution $P(\mathbf{p_i}), P(\mathbf{q_i})$}
	\While{not converged}
		\State{\emph{Adjust cluster proportions $\boldsymbol{\pi}$}}
		\State $\pi_k = \sum_{i=1}^m 1\{z_i = k\} / m$
		\For{$k=1,\dots,K$}
			\State{\emph{Adjust cluster hyperparameters $\boldsymbol{\alpha}_k, \boldsymbol{\beta}_k$}}
			\State $\boldsymbol{\alpha}_k^{new} = \argmax_{\boldsymbol{\alpha}_k} P(X,\Delta | \boldsymbol{\alpha}_k, z)$ \hfill \eqref{eq:adjust_alpha} \label{line:adjust_alpha}
			\State $\boldsymbol{\beta}_k^{new} = \argmax_{\boldsymbol{\beta}_k} P(X,\Delta | \boldsymbol{\beta}_k, z)$ \hfill \eqref{eq:adjust_beta} \label{line:adjust_beta}
		\EndFor
		\For{$i=1,\dots,m$}
			\State {\emph{Adjust users' assignments to clusters:}}
			\State $z_i^{new} = \argmax_{k} P(X, \Delta | z_i = k)$ \hfill \eqref{eq:cluster_assignment} \label{line:adjust_z}
		\EndFor
	\EndWhile
	\State \emph{Compute user posterior distributions:} \label{line:posterior}
	\State $P(\mathbf{p_i}|X,\Delta) = \text{Dirichlet}(\boldsymbol{\alpha}_{z_i} + \mathbf{n}^x_i)$ \label{line:ptilde}
	\State $P(\mathbf{q_i}|X,\Delta) = \text{Dirichlet}(\boldsymbol{\beta}_{z_i} + \mathbf{n}^\Delta_i)$
\EndProcedure
\end{algorithmic}
\end{algorithm}

\subsection{\textsc{NEST}: Proposed Metric for Detecting Suspicious Users}
\label{sec:metric}

Algorithm \ref{alg:fitting} gives us the posterior distributions $P(\mathbf{p_i} | x_i, \Delta_i)$ and $P(\mathbf{q_i} | x_i, \Delta_i)$ for the user parameters. In this section, we propose a suspiciousness metric, \metric (Normalized Expected Surprise Total). Recalling Figure \ref{fig:posterior_distributions}, the overall idea is to compute user $i$'s suspiciousness, averaged over their posterior distribution.

We will compute suspiciousness with respect to rating and temporally, then normalize and combine them to ensure that each has equal influence. This is a practically motivated decision that ensures that even in settings where one of the variables has a much finer resolution than the other (i.e. it is bucketized into more buckets), neither variable will dominate the other in determining suspiciousness. 

We now explain how to compute user $i$'s suspiciousness is in terms of their ratings distribution; the same formulas directly apply to the temporal distribution, and we explain how to combine the scores in \eqref{eq:nest}. 

\subsubsection*{Global Distribution} Recall that our Bayesian model \model gives us an estimate for the distribution underlying the rating behavior of all users, in the form of a mixture of Dirichlet($\boldsymbol{\alpha}_k$) distributions with mixture coefficients $\boldsymbol{\pi}_k$ (and similarly, mixture of Dirichlet($\boldsymbol{\beta}_k$) distributions for temporal distributions). Denote this global distribution by $F_x$ (resp. $F_\Delta$):

$F_x$ can be thought of as our estimate for the distribution of $\mathbf{p_i}$ in general over all users ($\mathbf{p_i}$ is the true rating distribution of user $i$).
\begin{align}
F_x(\mathbf{p}) = \sum_{k=1}^K \boldsymbol{\pi}_k \ \text{Dirichlet}(\mathbf{p};\boldsymbol{\alpha}_k)
\label{eq:global_distribution}
\end{align}
where $\text{Dirichlet}(\mathbf{p};\boldsymbol{\alpha}_k)$ refers to the probability of generating $\mathbf{p}$ under a $\text{Dirichlet}(\boldsymbol{\alpha}_k)$ distribution.

\subsubsection*{Surprise} 
Denote $\tilde{\mathbf{p_i}} := P(\mathbf{p_i} | x_i, \Delta_i)$, the posterior distribution of $\mathbf{p_i}$ given the data. To be clear, observe that $\tilde{\mathbf{p_i}}$ is a distribution over multinomial vectors. Recall that we estimate $\tilde{\mathbf{p_i}}$ as part of \model: $\tilde{\mathbf{p_i}}$ is a $\text{Dirichlet}(\boldsymbol{\alpha}_{z_i} + \mathbf{n}^x_i)$ distribution. $\tilde{\mathbf{p_i}}$ represents our beliefs about user $i$'s rating distribution. 

For the sake of intuition, imagine $\tilde{\mathbf{p_i}}$ was a point mass, i.e. we had perfect knowledge of user $i$'s rating distribution: assume that it consists of a point mass at $\mathbf{p}$. Recall that the posterior distribution $\tilde{\mathbf{p_i}}$ is a distribution over multinomial vectors, so $\mathbf{p}$ here is a multinomial vector. Then user $i$'s suspiciousness could be calculated as \emph{surprise} or negative log likelihood under the global distribution $F_x$ evaluated at the rating distribution $\mathbf{p}$:
\[\text{surprise}(\mathbf{p}) = -\log F_x(\mathbf{p})\]
The less likely $\mathbf{p}$ is, the more suspicious the user is. This makes sense because $F_x$ is our estimate for the global distribution from which all the users are drawn from; the lower the log-likelihood of $\mathbf{p}$, the more anomalous user $i$ is when compared to this distribution. Thus, we use \emph{surprise} to estimate the suspiciousness of a rating distribution $\mathbf{p}$. 

\subsubsection*{Expected Surprise} Now returning to the general case, when $\tilde{\mathbf{p_i}}$ is a posterior distribution. In this case, we compute the average over $\tilde{\mathbf{p_i}}$ of the \emph{surprise} $-\log F_x(\mathbf{p})$: that is, now $\mathbf{p}$ is drawn at random from this posterior distribution $\tilde{\mathbf{p_i}}$. Averaging the \emph{surprise} gives us the posterior mean (or `Bayes estimate') of user $i$'s suspiciousness, which can be regarded as our `best estimate' of user $i$'s suspiciousness given our knowledge of them.\footnote{The posterior mean of a parameter is also known as the minimum mean square error estimator, as it minimizes expected least squares loss. It has desirable properties such as consistency under fairly general conditions, and is widely used in practice.\cite{lehmann1998theory}} Thus, we use \emph{expected surprise} to estimate the suspiciousness of a user based on their posterior distribution $\tilde{\mathbf{p_i}}$. 

\begin{definition}[Expected Surprise] \hfill \\
The expected surprise for user $i$ measures how surprising user $i$ rating distribution is averaged over its posterior distribution, and is given by:
\begin{align}
 s_x(i) = -\mathbb{E}_{\mathbf{p} \sim \tilde{\mathbf{p_i}}} \log F_x(\mathbf{p})
 \label{eq:expected_surprise}
 \end{align}
\end{definition}
The expected surprise $s_\Delta(i)$ with respect to the temporal distribution is computed similarly. 

As discussed, $-\mathbb{E}_{\mathbf{p} \sim \tilde{\mathbf{p_i}}} \log F_x(\mathbf{p})$ measures the expected suspiciousness of a sample rating distribution drawn at random from the posterior distribution $\tilde{\mathbf{p_i}}$, where suspiciousness of a single rating distribution is given by its \emph{surprise} or negative log likelihood under $F_x$.

\subsubsection*{Normalized Expected Surprise Total (\metric)} 
In our dataset, we use both ratings and temporal data. Using ratings data, we compute posterior distribution $\tilde{\mathbf{p_i}}$ and the resulting \emph{expected surprise} $s_x(i)$; using temporal data similarly gives us posterior distribution $\tilde{\mathbf{q_i}}$ and \emph{expected surprise} $s_\Delta(i)$. To combine these, we could simply add them; however, if one had a larger range of possible values than the other, the one with the largest range could end up dominating the sum. To give both terms comparable influence, we normalize them by their respective standard deviations. Let $\sigma_x = \text{std.dev}(s_x(1), \dots, s_x(m))$ and $\sigma_\Delta$ be defined analogously. Then \metric is defined as:

\begin{definition}[\metric] \hfill \\
\metric measures how jointly suspicious user $i$ is based on his or her ratings and temporally, and is given by:
\begin{align}
 \textsc{NEST}(i) = \frac{s_x(i)}{\sigma_x} + \frac{s_\Delta(i)}{\sigma_\Delta}
 \label{eq:nest}
 \end{align}
\end{definition}

Note that there is no need to normalize $s_x$ and $s_\Delta$ additively (e.g. by subtracting the mean scores) since that would simply shift all the scores by the same amount. 

\subsubsection*{Computing \metric}
Fitting \model gives us the posterior distribution for user $i$'s rating distribution $\tilde{\mathbf{p_i}}=P(\mathbf{p_i}|X,\Delta)$ (Line \ref{line:ptilde} in Algorithm \ref{alg:fitting}); we get $\tilde{\mathbf{q_i}}=P(\mathbf{q_i}|X,\Delta)$ similarly. We also know the full global distribution $F_x(\mathbf{p}) = \sum_{k=1}^K \boldsymbol{\pi}_k \ \text{Dirichlet}(\mathbf{p};\boldsymbol{\alpha}_k)$. Hence, we can compute the expected surprise $s_x(i) = -\mathbb{E}_{\mathbf{p} \sim \tilde{\mathbf{p_i}}} \log F_x(\mathbf{p})$ by taking a fixed number of samples from $\tilde{\mathbf{p_i}}$ and repeatedly evaluating their log-likelihood under $F_x$. We can then compute the expected surprise values $s_\Delta(\cdot)$ and combining the two as shown in \eqref{eq:nest}.

\label{sec:Results}
\section{Experiments}

We conducted experiments to answer the following questions: \textbf{Q1. Effectiveness on real data}: does \method catch fraud on real data? \textbf{Q2. Scalability}: does it scale to large datasets? \textbf{Q3: Interpretability}: can the results of the Bayesian model \model and the scores given by \metric be interpreted in a real-life setting?

We implemented \method in Python; all experiments were carried out on a 2.4 GHz Intel Core i5 Macbook Pro, 16 GB RAM, running OS X 10.9.5. The code is available for download at \codeurl. We test \method on a variety of real world datasets: table \ref{tab:datasets} offers details on the datasets we used.

\begin{table}[htp]
\centering
\caption{Datasets used.}
\label{tab:datasets}
\begin{tabular}{@{}lrll@{}}
\toprule
\textbf{Dataset} & \textbf{$\#$ of users} & \textbf{$\#$ of products} & \textbf{$\#$ of ratings} \\ \midrule
\texttt{Flipkart} &  1.1M & 550K & 3.3M \\ 
\texttt{SWM}\cite{akoglu2013opinion} & 0.97M & 15K & 1.1M\\  \bottomrule
\end{tabular}
\end{table}

\subsection{Q1: Effectiveness}

\subsubsection*{Evaluation on \texttt{Flipkart} data}

\texttt{Flipkart} is an online e-commerce platform on which merchants sell products to customers, on which customers review products from $1$ to $5$ stars. We applied \method to detect the $250$ most suspicious users and provided them to Flipkart; these accounts were investigated and hand-labelled by Flipkart, finding that $211$ users of the top $250$ flagged by \method were involved in fraud. Figure \ref{fig:flipkart_pak} shows the algorithm's precision at $k$: for various values of $k$ up to $250$: note that precision for the most suspicious users is very high: e.g. precision of $1.0$ for the first $50$ users. These are substantial findings for Flipkart. One common pattern that the domain-experts found was that most of the users labeled as fraudulent are either spamming 4/5 star ratings to multiple products from a single seller (boosting seller's ratings), or spamming 1/2 star ratings to multiple products from another seller (defaming the competition).

Figure \ref{fig:flipkart_ratings_goodbad} plots the averaged rating distributions of users within each group: that is, for each user we computed their frequency of giving each rating from $1$ to $5$;  Figures \ref{fig:flipkart_ratings_goodbad} and \ref{fig:flipkart_iat_goodbad} takes the average of the rating distributions for users in the corresponding group. Examining the detected users in Figure \ref{fig:flipkart_ratings_goodbad}, a common pattern we find is that they consist of extreme polarized rating distributions as well as temporal distributions. The detected users consist of highly negative users (who give only $1$ ratings) and highly positive users (who give mostly $5$ ratings, with a relatively small fraction of $4$s). Similarly, Figure \ref{fig:flipkart_iat_goodbad} shows the common pattern that detected users contain much shorter temporal differences than normal users. 

\subsubsection*{Evaluation on \texttt{SWM} data}

The \texttt{SWM} datasets consists of software product (app) reviews. The dataset was collected by \cite{akoglu2013opinion} by crawling all the app reviews in the entertainment category from an anonymous online app store. Each review consists of review text as well as a rating from $1$ star to $5$ stars. 

We find clear evidence of fake reviews in the dataset: for example, the most suspicious user posted a block of $27$ reviews for the same app all within the span of less than a week, all five-star ratings with near-identical review title and text, as shown in Table \ref{tab:app_review}. Moreover, the review text shows clear signs of being a fake review: they advertise a code associated with an app: typically, users advertise such codes because they give some benefit to the owner when new users download the app via one of these codes. 

\begin{table}[htp]
\caption{\textbf{\method detects clearly fake reviews in the \texttt{SWM} data:} Example of a $5$-star review by the user flagged as most suspicious by \method. $27$ such near-identical reviews were present for the same app (only trivial differences between them were present, such as the number of dollars signs). All $10$ of the top $10$ user accounts flagged by \method contain similar advertisements for codes.}
\centering
\label{tab:app_review}
\begin{tabular}{|l|}
\hline
\textbf{\textit{AWESOMEApp4FreeMoney!!! \$\$\$\$\$\$}}  \\ \hline
\emph{All first time users will need a} \\
\emph{CODE after downloading this app. So} \\
\emph{download it now and use my CODE for} \\
\emph{bonus points. CODE: ...} \\
\hline
\end{tabular}
\end{table}

Aside from this block of reviews, almost all of this user's reviews also consist of similar blocks of repeated text advertising the code. In fact, all $10$ of the top suspicious user accounts flagged by \method contain advertisements for codes in similar contexts, often accompanied with promises of free cash, points and gift cards. 

\subsection{Q2: Scalability}

\begin{figure}[htp]
    \centering
    \includegraphics[width=0.25\textwidth]{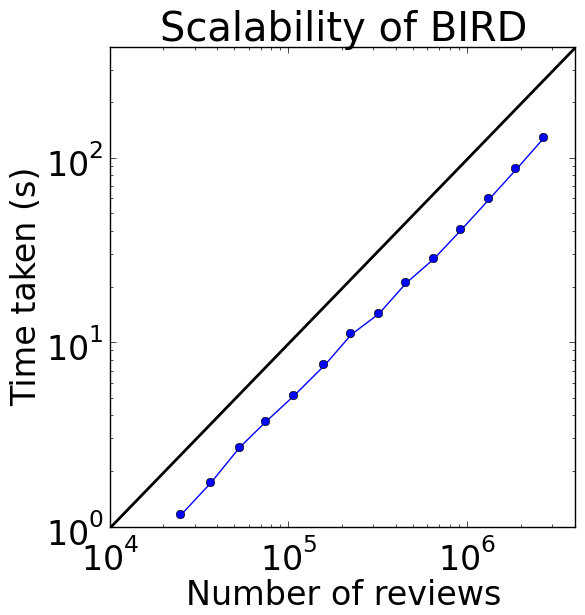}
    \caption{\textbf{\textsc{BIRDNEST} is fast and scalable:} running on our 1.1M user, 3.3M ratings dataset in around 2 minutes. \method shows linear growth in computation time.}
\label{fig:timing}
\end{figure}

Assume that we run $(\#it)$ iterations of the outer loop of Algorithm \ref{alg:fitting}, let $m$ be the number of users, and $K$ the number of clusters. In each iteration, adjusting $\boldsymbol{\pi}$ takes $O(m)$ time; adjusting each $\boldsymbol{\alpha}_i$, $\boldsymbol{\beta}_i$ and $z_i$ all take $O(K)$ time. Hence, the algorithm takes $O((\#it)mK)$ time, which is linear. Figure \ref{fig:timing} shows that the algorithm is fast and its computation time grows linearly in practice. 

\subsection{Q3: Interpretability}

In Section \ref{sec:motivating}, we motivated the Bayesian approach by giving three users, Alice, Bob and Carol. We explained how the Bayesian approach captures our intuitions about these users through posterior distributions. We now use real data from Flipkart to verify that \method indeed conforms to the intuitions that motivated this approach, and that the posterior distributions from \method are interpretable and useful in real-life settings. 

\begin{figure}[htp]
    \centering
   \includegraphics[width=0.35\textwidth]{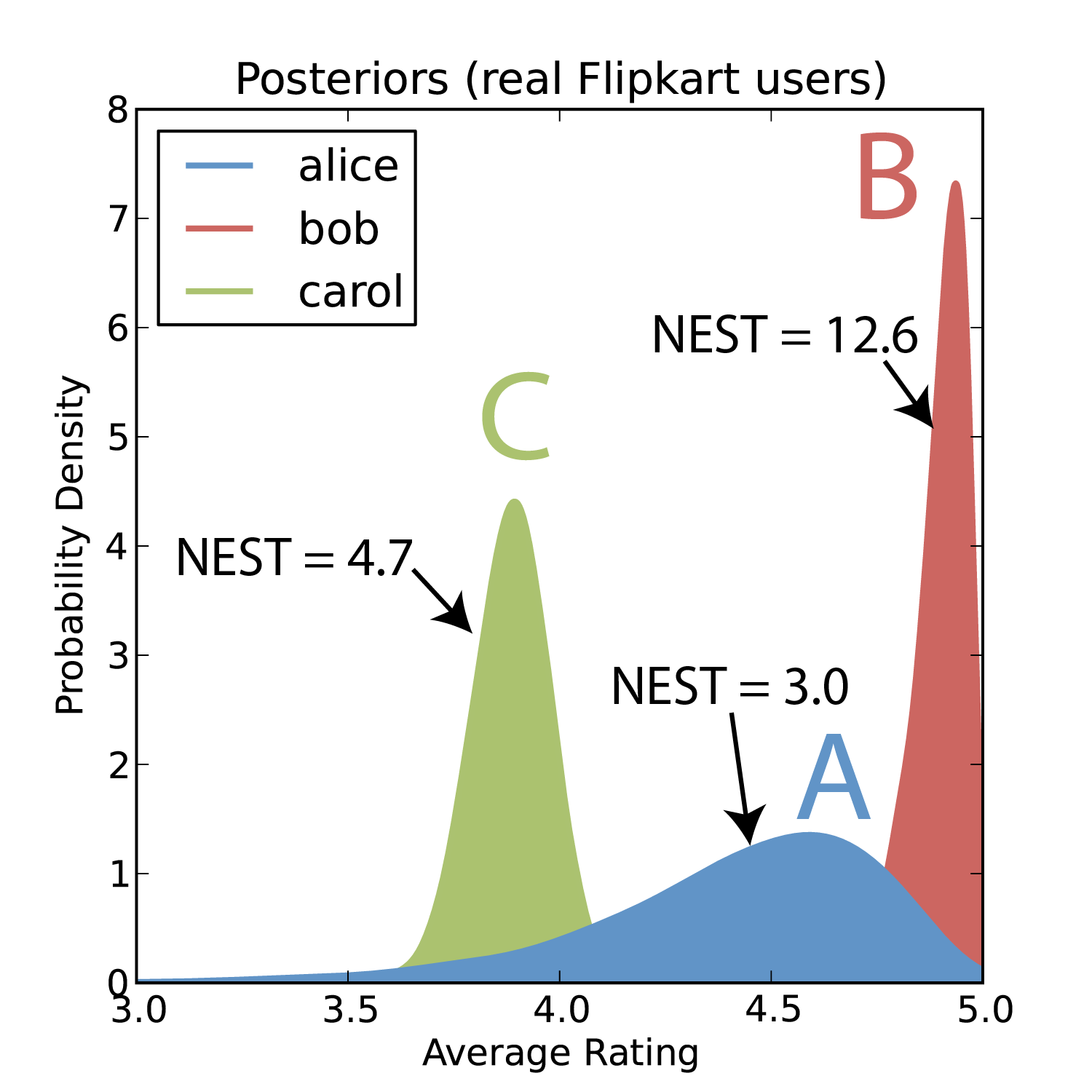}
    \caption{\textbf{\textsc{BIRDNEST} is interpretable and agrees with intuition: } the posterior distributions for users capture both the location and certainty about a user's true rating distribution. Note that \texttt{bob}'s \metric is highest as his entire posterior distribution is extreme (far from other users). }
	\label{fig:flipkart_abc}
\end{figure}

We selected $3$ real Flipkart users: \texttt{alice}, \texttt{bob} and \texttt{carol} (names changed to maintain anonymity). They were chosen to match the rating frequencies of Alice, Bob and Carol in \ref{sec:motivating}. We computed each user's posterior distribution of their true rating distribution using \method. From this we computed the posterior distribution of their true, long-term average rating, by simulating $10,000$ draws of $\mathbf{p_i}$ from their Dirichlet posterior distribution. We display these with their \metric values in Figure \ref{fig:flipkart_abc}. 

Agreeing with intuition, both \texttt{alice} and \texttt{carol} are nonsuspicious, while \texttt{bob} is very suspicious, as indicated by the \metric scores: \texttt{alice} is ranked around $189,000$th most suspicious, \texttt{bob} is ranked around $800$th, and \texttt{carol} is ranked around $10,000$th, out of the $1.1$ million users. The \metric scores given by our algorithm are interpretable as \emph{expected surprise} values, i.e. they are in units of log-likelihood, so a unit difference (after normalization) represents an exponential increase in likelihood. As such, we see from this example that \method conforms to intuition in the way it uses posterior distributions to measure uncertainty. Moreover, the posterior distributions of average rating or other quantities can be plotted via simulation and used for further understanding and investigation.

\section{Conclusion}
\label{sec:Conclusion}
In this paper, we developed \model, a Bayesian inference approach for ratings data, and \metric, a principled likelihood-based suspiciousness metric for fraud detection. Our method provides a principled way to combine rating and temporal information to detect rating fraud, and to find a tradeoff between users with extreme rating distributions vs. users with larger number of ratings. Our contributions are:
\bit
\item \textbf{Theoretically sound user behavior model}: we define a Bayesian model for the data based on a mixture model which captures different types of user behavior. This model then allows us to determine how much an anomalous user deviates from normal behavior.
\item \textbf{Suspiciousness metric}: we define a likelihood-based metric which measures how much a user deviates from normal behavior.
\item \textbf{Algorithm}: we propose a scalable and effective algorithm for learning the Bayesian model and evaluating suspiciousness.
\item \textbf{Effectiveness}: we show that our method successfully spots review fraud in large, real-world graphs, with precision of over $84\%$ for the top $250$ Flipkart users flagged by our algorithm.
\eit

\bibliographystyle{abbrv}
\bibliography{rating}

\end{document}